\documentclass[letterpaper, 10 pt, conference]{ieeeconf}  

\IEEEoverridecommandlockouts                              

\usepackage{graphics} 
\usepackage{epsfig} 
\usepackage{amsmath} 
\usepackage{amssymb}  
\usepackage{balance}
\usepackage{amsfonts}
\usepackage{booktabs} 
\usepackage{cite}
\usepackage{makecell}
\usepackage[backref]{hyperref}
\hypersetup{
hidelinks,
colorlinks=true,
linkcolor=black,
urlcolor=black,
citecolor=black
}
\usepackage{color}
\usepackage{soul}
\sethlcolor{yellow}
\graphicspath{{figures/}}
\usepackage{rotating}
\usepackage{pifont}
\usepackage{colortbl}
\usepackage{array} 
\usepackage{multirow}
\usepackage{graphicx}
\usepackage[table,xcdraw]{xcolor}

\newcommand{\cmark}{\ding{51}} 
\newcommand{\xmark}{\ding{55}} 
\newcommand{\first}[1]{\textcolor{red}{\textbf{#1}}}
\newcommand{\second}[1]{\textcolor{blue}{\textbf{#1}}}

\definecolor{barriercolor}{RGB}{255, 120, 50}
\definecolor{bicyclecolor}{RGB}{255, 192, 203}
\definecolor{buscolor}{RGB}{255, 255, 0}
\definecolor{carcolor}{RGB}{0, 150, 245}
\definecolor{constructcolor}{RGB}{0, 255, 255}
\definecolor{motorcolor}{RGB}{200, 180, 0}
\definecolor{pedestriancolor}{RGB}{255, 0, 0}
\definecolor{trafficcolor}{RGB}{255, 240, 150}
\definecolor{trailercolor}{RGB}{135, 60, 0}
\definecolor{truckcolor}{RGB}{160, 32, 240}
\definecolor{drivablecolor}{RGB}{255, 0, 255}
\definecolor{otherflatcolor}{RGB}{139, 137, 137}
\definecolor{sidewalkcolor}{RGB}{75, 0, 75}
\definecolor{terraincolor}{RGB}{150, 240, 80}
\definecolor{manmadecolor}{RGB}{230, 230, 255}
\definecolor{vegetationcolor}{RGB}{0, 175, 0}
\definecolor{otherscolor}{RGB}{0, 0, 0}

\title{\LARGE \bf
Deep Height Decoupling for Precise Vision-based \\ 3D Occupancy Prediction 
}

\author{Yuan Wu$^{1}$$^{\ast}$, Zhiqiang Yan$^{1}$$^{\ast}$, Zhengxue Wang$^{1}$, Xiang Li$^{2}$, Le Hui$^{3}$, and Jian Yang$^{1}$$^{\dagger}$
    \thanks{$^{1}$PCA Lab, Key Lab of Intelligent Perception and Systems for High-Dimensional Information of Ministry of Education, and Jiangsu Key Lab of Image and Video Understanding for Social Security, School of Computer Science and Engineering, Nanjing University of Science and Technology. {\tt\small \{wuyuan,yanzq,zxwang,csjyang\}@njust.edu.cn}}%
    \thanks{$^{2}$College of Computer Science, Nankai University, Tianjin 300071, China. {\tt\small xiang.li.implus@nankai.edu.cn}}%
    \thanks{$^{3}$Electronics and Information, Northwestern Polytechnical University, Xi'an 710072, China. {\tt\small huile@nwpu.edu.cn}}%
    \thanks{$^{\ast}$Equal contribution}%
    \thanks{$^{\dagger}$Corresponding author}%
}

\begin{document}

\maketitle
\thispagestyle{empty}
\pagestyle{empty}

\begin{abstract}
%
The task of vision-based 3D occupancy prediction aims to reconstruct 3D geometry and estimate its semantic classes from 2D color images, where the 2D-to-3D view transformation is an indispensable step. 
%
%
Most previous methods conduct forward projection, such as BEVPooling and VoxelPooling, both of which map the 2D image features into 3D grids. 
%
However, the current grid 
representing features within a certain height range usually introduces many confusing features that belong to other height ranges. 
%
To address this challenge, we present Deep Height Decoupling (DHD), a novel framework that incorporates explicit height prior to filter out the confusing features. 
Specifically, DHD first predicts height maps via explicit supervision. Based on the height distribution statistics, DHD designs Mask Guided Height Sampling (MGHS) to adaptively decouple the height map into multiple binary masks. 
MGHS projects the 2D image features into multiple subspaces, where each grid contains features within reasonable height ranges. 
Finally, a Synergistic Feature Aggregation (SFA) module is deployed to enhance the feature representation through channel and spatial affinities, enabling further occupancy refinement. 
On the popular Occ3D-nuScenes benchmark, our method achieves state-of-the-art performance even with minimal input frames. 
Source code is released at \url{https://github.com/yanzq95/DHD}. 
\end{abstract}

\section{Introduction}
Understanding the 3D geometry and semantic information of surrounding environment is crucial for autonomous driving. In recent years, the rapid development of camera-based algorithms has significantly advanced outdoor 3D scene understanding \cite{huang2021bevdet,li2022bevformer,CGFormer,huang2024gaussianformer,peng2024learning,yan2024tri,wang2024dcdepth,yan2022rignet,zhu2023curricular,zheng2023occworld}, leading to increased attention on the vision-based 3D occupancy prediction task \cite{li2023voxformer,wei2023surroundocc,tian2024occ3d,gan2024gaussianocc,shi2024occupancysetpoints,li2024viewformer}.  

\begin{figure}[t]
    \centering
    \includegraphics[width=8.5cm]{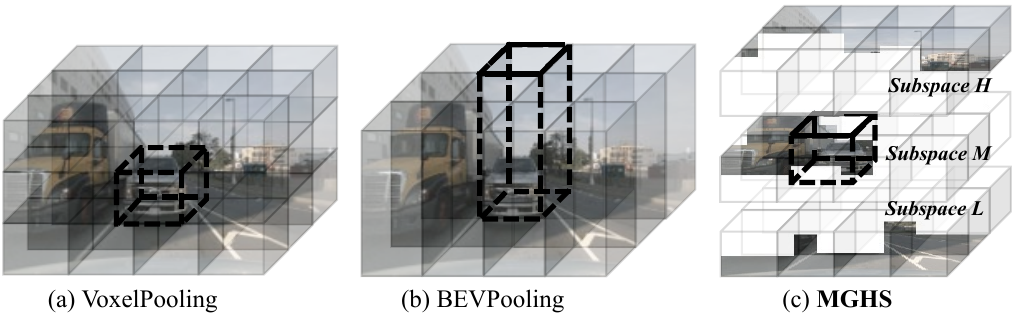}
    \caption{Projection comparison. (a) VoxelPooling \cite{huang2021bevdet,li2023bevstereo} retains height but overlooks class-specific height distributions. (b) BEVPooling \cite{yu2023flashocc,yu2024panoptic} sacrifices height details by collapsing the height dimension. In contrast, (c) our mask guided height sampling (MGHS) selectively projects 2D features based on object heights, preserving more accurate and detailed features.}
    \label{fig:BEVPool_VoxelPool_Ours}
\end{figure}

This task estimates 3D occupancy from 2D color images. A fundamental step in this process is the 2D-to-3D view transformation. Traditional methods \cite{li2023bevdepth,li2023bevstereo,hou2024fastocc,yu2023flashocc,yu2024panoptic,myeongjin2023milo} utilize depth information \cite{yan2023distortion,yan2022multi} to map the 2D image features into 3D grids. For example, the VoxelPooling in Fig.~\ref{fig:BEVPool_VoxelPool_Ours} (a) divides 3D space into fixed-size voxels and aggregates features in each grid, while the BEVPooling in Fig.~\ref{fig:BEVPool_VoxelPool_Ours} (b) accumulates all frustum features in the flattened BEV grid. That is to say, each of these grids is likely to predict the objects whose height ranges do not belong to the current grid. It is challenging for networks to handle such situations. 

\begin{figure}[t]
    \centering
    \includegraphics[width=8.6cm]{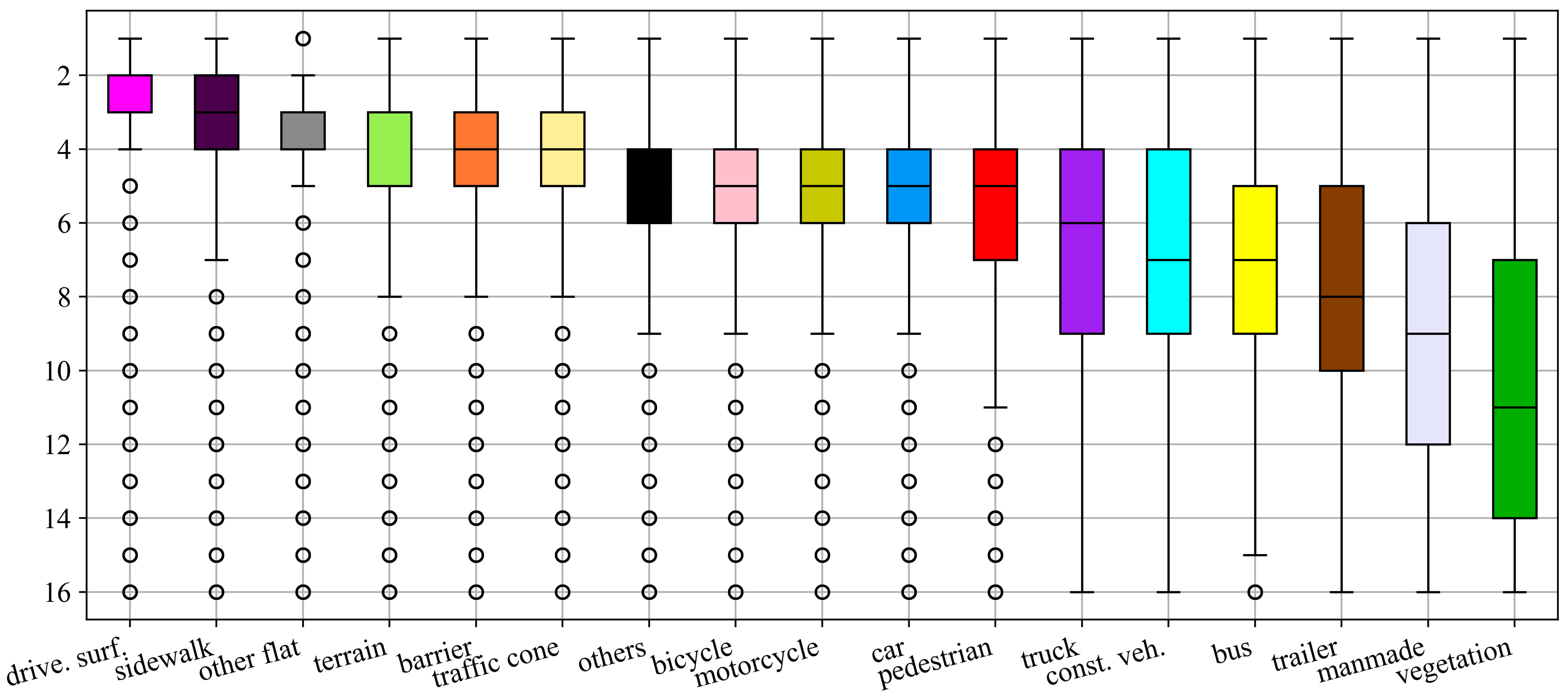}
    \caption{Height distribution of different classes on Occ3D-nuScenes \cite{tian2024occ3d}.}
    \vspace{-4pt}
    \label{fig:box}
\end{figure}

Moreover, as illustrated in Fig.~\ref{fig:box}, the statistics of the entire Occ3D-nuScenes dataset indicate that different objects usually possess varying height distributions. For instance, the \emph{manmade} class occupies larger height ranges, while the \emph{car} and \emph{pedestrian} classes are confined to lower height ranges. Consequently, both the VoxelPooling and BEVPooling approaches encounter challenges due to the introduction of confusing features from different height ranges.

\begin{figure*}[t]
    \centering
    \includegraphics[width=17.5cm]{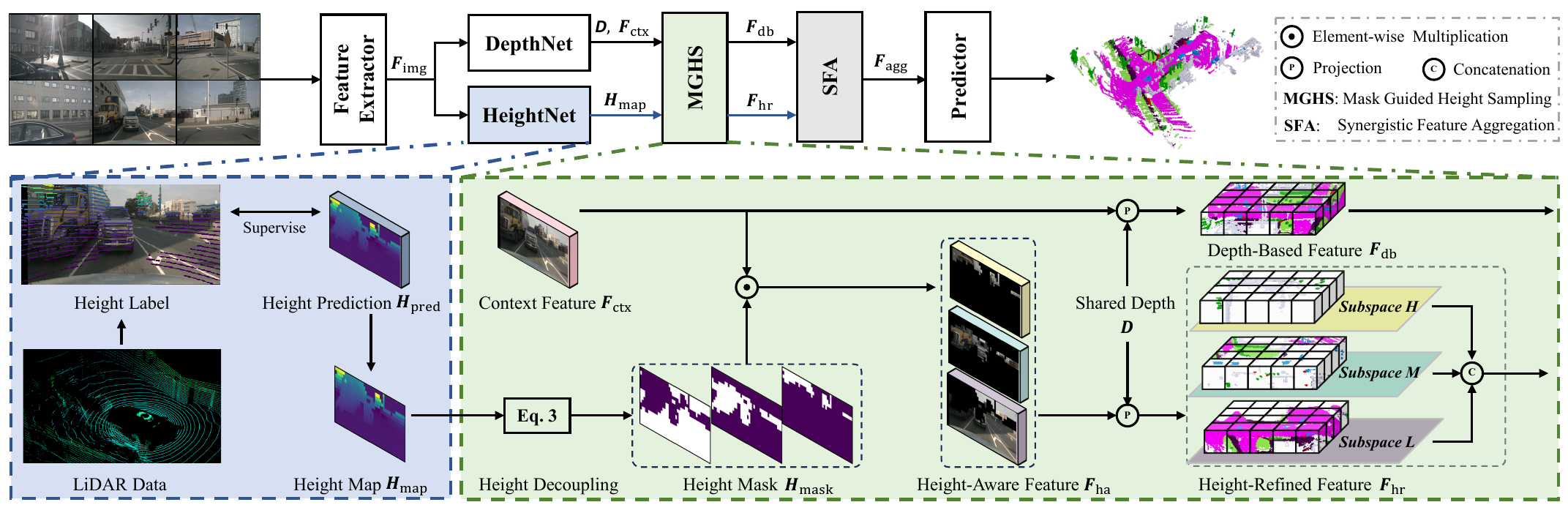}
    \vspace{-8pt}
    \caption{An overview of our deep height decoupling (DHD) framework (see section \ref{sec:DHD} for details).}
    \vspace{-5pt}
    \label{fig:model}
\end{figure*}

To address these challenges, we propose a novel framework called deep height decoupling (DHD). 
For the first time, DHD introduces explicit height prior to decouple the 3D features and filter out the redundant parts. Specifically, DHD first generates height maps through LiDAR supervisory signals, capturing the height distribution of the scene. Then, to achieve precise and effective 2D-to-3D view transformation, DHD designs a mask guided height sampling (MGHS) module to sample features across varying heights. Based on reasonable height distribution statistics, MGHS decouples the height maps into multiple height masks, each corresponding to a distinct height range. These masks are applied to filter out the 2D features for height-aware feature sampling. Finally, the masked 2D features are projected into multiple 3D subspaces, ensuring that the 3D features within each grid fall within reasonable height ranges. Additionally, DHD introduces a synergistic feature aggregation (SFA) module, which enhances the dual feature representations by leveraging both channel and spatial affinities, contributing to further occupancy refinement. 

In summary, our contributions are as follows: 
\begin{itemize}
    \item We present the DHD framework for occupancy prediction, which, for the first time, incorporates the height prior into the model through explicit height supervision. 
    \item We propose MGHS with height decoupling and sampling, enabling precise feature projection. It ameliorates the problem of feature confusion in the 2D-to-3D view transformation, a crucial step of all occupancy models. 
    \item We introduce SFA to enhance the feature representation for further occupancy refinement. 
    \item DHD achieves superior performance with minimal input cost\footnote{Using only one history frame in temporal fusion.}. Codes are released for peer research. 
\end{itemize}

\section{Related Work}
\subsection{Vision-Based 3D Occupancy Prediction}
Recently, vision-based 3D occupancy prediction \cite{tian2024occ3d,wei2023surroundocc,gan2024gaussianocc} has garnered increasing attention, including both supervised and unsupervised methods. For supervised methods~\cite{huang2021bevdet,cao2022monoscene,li2022bevformer,li2023voxformer}, MonoScene \cite{cao2022monoscene} is the pioneering approach for monocular occupancy prediction.
BEVDet \cite{huang2021bevdet} employs Lift-Splat-Shoot (LSS) \cite{philion2020lift} to project 2D image features into 3D space. 
BEVDet4D \cite{huang2022bevdet4d} further explores the temporal fusion strategy by fusing features from the previous frames.
Besides, several transformer-based methods \cite{li2022bevformer,li2023voxformer,liu2023fully} have been proposed. 
For example, BEVformer \cite{li2022bevformer} utilizes spatiotemporal transformers to construct BEV features. 
VoxFormer~\cite{li2023voxformer} introduces a sparse-to-dense transformer framework for 3D semantic scene completion.
For unsupervised methods~\cite{huang2024selfocc,zhang2023occnerf,pan2023renderocc}, SelfOcc~\cite{huang2024selfocc} and OccNeRF~\cite{zhang2023occnerf} are two representative works that employ volume rendering and photometric consistency to generate self-supervised signals. 
Different from previous methods that directly extract 3D features, we introduce height prior to decouple height, effectively capturing finer details across different height ranges.

\subsection{2D-to-3D View Transformation}
Many existing methods leverage depth information \cite{yan2023desnet,yan2024learnable,wang2021regularizing,yan2024completion,yan2023rignet++} to perform 2D-to-3D view transformation. LSS \cite{philion2020lift} explicitly predicts the depth distribution and elevates 2D image features into 3D space.
Most recently, a few methods~\cite{li2023bevdepth,li2023bevstereo,miao2023occdepth}
emphasize the significance of accurate depth estimation in view transformation.  
For example, BEVDepth \cite{li2023bevdepth} encodes camera intrinsics and extrinsics into a depth refinement module for 3D object detection. BEVStereo \cite{li2023bevstereo} introduces an effective temporal stereo technique to enhance depth estimation.
Additionally, some researches~\cite{zhang2023sa,xie2023sparsefusion,huang2022bevpoolv2,chi2023bev} focus on optimizing the projection stage. 
For instance, SparseFusion \cite{xie2023sparsefusion} argues that projecting all virtual points into the BEV space is unnecessary. 
Similarly, SA-BEV \cite{zhang2023sa} utilizes SA-BEVPool to project only foreground image features. 
Differently, we employ height masks to selectively project features, enhancing the accuracy of spatial representation.

\section{Deep Height Decoupling} 
\label{sec:DHD}
\subsection{Overview}
\label{sub:overview}
As illustrated in Fig.~\ref{fig:model}, our DHD comprises a feature extractor, HeightNet, DepthNet, MGHS, SFA, and predictor. The feature extractor first acquires 2D image feature $\boldsymbol{F}_\mathrm{img}\in \mathbb{R}^{n\times c\times h\times w}$ from $n$ cameras.
Then, DepthNet extracts context feature $\boldsymbol{F}_\mathrm{ctx}$ and depth prediction $\boldsymbol{D}$. HeightNet generates the height map $\boldsymbol{H}_\mathrm{map}$ to determine the height value at each pixel.
Next, MGHS integrates $\boldsymbol{D}$, $\boldsymbol{F}_\mathrm{ctx}$, and $\boldsymbol{H}_\mathrm{map}$ for feature projection. 
Specifically, it decouples $\boldsymbol{H}_\mathrm{map}$ to produce height masks $\boldsymbol{H}_\mathrm{mask}$ for different height ranges, which are utilized to filter the height-aware feature $\boldsymbol{F}_\mathrm{ha}$ from $\boldsymbol{F}_\mathrm{ctx}$. 
$\boldsymbol{F}_\mathrm{ha}$ is projected into multiple subspaces to obtain height-refined feature $\boldsymbol{F}_\mathrm{hr}$. 
Besides it employs BEVPooling to encode $\boldsymbol{F}_\mathrm{ctx}$ into depth-based feature $\boldsymbol{F}_\mathrm{db}$. 
Finally, both $\boldsymbol{F}_\mathrm{hr}$ and $\boldsymbol{F}_\mathrm{db}$ are fed into the SFA to obtain the aggregated feature $\boldsymbol{F}_\mathrm{agg}$, which serves as input for the predictor.

\begin{figure*}[t]
    \centering
    \includegraphics[width=17.4cm]{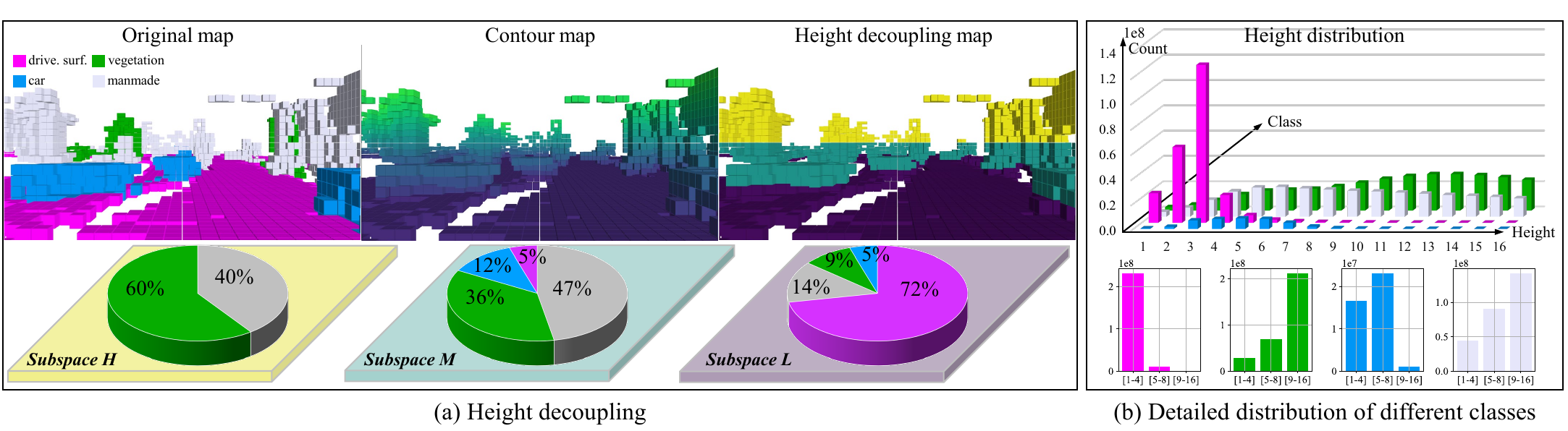}
    \vspace{-9pt}
    \caption{(a) We decouple height into three intervals to differentiate features across heights and list the proportion of each class below. 
    (b) The distribution of various classes across different heights, with the bar chart presenting statistical values within each interval. 
    }
    \label{fig:Height_Sep3_with_statics}
    \vspace{-10pt}
\end{figure*}
\begin{figure}[t]
    \centering
    \includegraphics[width=8.5cm]{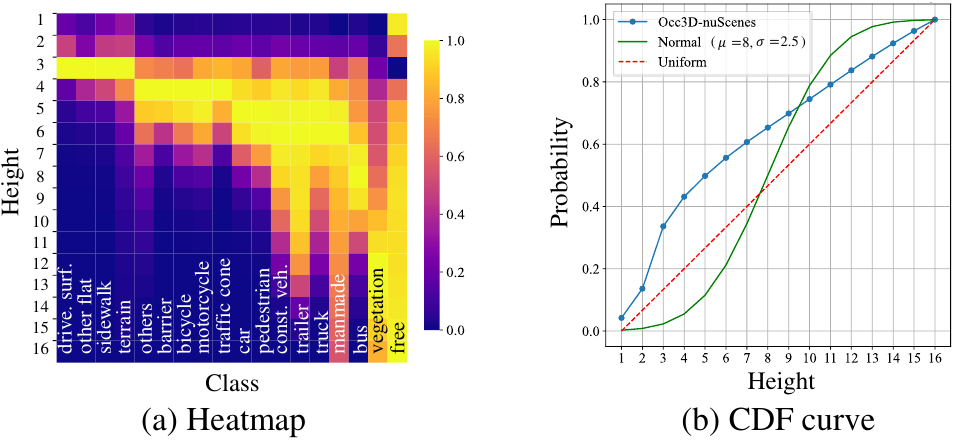}
    \vspace{-8pt}
    \caption{Semantic and geometric analysis of the Occ3D-nuScenes \cite{tian2024occ3d}. (a) Heatmap illustrates the normalized distribution of each category across different heights. (b) Cumulative distribution function (CDF) curve suggests the data concentrates in specific height layers.
    }
    \vspace{-8.5pt}
    \label{fig:heatmap_CDF}
\end{figure}

\subsection{HeightNet}
\label{sub:Explicit Height Supervision}
Inspired by the DepthNet in BEVDepth \cite{li2023bevdepth}, we reformulate the regression task by employing one-hot encoding, discretizing the height into bins.
Specifically, we first utilize the SE-layer \cite{hu2018squeeze} and deformable convolution~\cite{zhu2019deformable} to predict height $\boldsymbol{H}_\mathrm{pred}$.
Then, the height value for each pixel is obtained by applying the argmax operation along the channel dimension, resulting in $\boldsymbol{H}_\mathrm{map}$.

Furthermore, we propose using ground-truth $\boldsymbol{H}_\mathrm{gt}$ from LiDAR points to supervise the learning of $\boldsymbol{H}_\mathrm{pred}$, facilitating the accurate height estimation. 
To obtain $\boldsymbol{H}_\mathrm{gt}$, given a 3D point $\boldsymbol{p}_{\mathrm{l}}=[x_\mathrm{l},y_\mathrm{l},z_\mathrm{l},1]^\top$ in the LiDAR coordinate system, we calculate its translate position $\boldsymbol{p}_{\mathrm{e}}=[x_\mathrm{e},y_\mathrm{e},z_\mathrm{e},1]^\top$  in the ego coordinate system.

Next, the 3D LiDAR points are projected onto the 2D image plane using perspective projection:
\begin{equation} d\left[u, v, 1\right]^\top = \boldsymbol{K} \left[\boldsymbol{R}',\ \boldsymbol{t}'\right] \left[x_\mathrm{l},\ y_\mathrm{l},\ z_\mathrm{l},\ 1\right]^\top, \label{Eq:2} \end{equation}
where $d$ is the depth value in the camera coordinate system, and $[u,v,1]^\top$ denotes the homogeneous representation of a point in the pixel coordinate system.  
$\boldsymbol{K}$ refers to the camera intrinsic. $\boldsymbol{R}'\in\mathbb{R}^{3\times3}$ and $\boldsymbol{t}'\in\mathbb{R}^{3}$ stand for the rotation and translation matrices from the LiDAR coordinate system to the camera coordinate system, respectively. 

As a result, we obtain a set of points represented as $[u,v,d,z_{\mathrm{e}}]^\top$, where each point is described by its pixel coordinates
$[u,v]^\top$, depth $d$, and height $z_{\mathrm{e}}$. Finally, we retain only the point with the smallest depth from those with identical pixel coordinates, yielding $\boldsymbol{H}_\mathrm{gt}$.

\begin{table}[t]
\caption{Entropy of different decoupling strategies. The gray background denotes the default setting.}
\label{tab:entropy_results}
\centering
\begin{tabular}{lc|c}
\toprule
Decoupling & Number & Entropy ($\times10^{-1}$)$\downarrow$\\
\midrule
\texttt{[}1, 16\texttt{]}                                                    & 1 & 4.69 ($\pm$0.00)\\ 
\midrule
\texttt{[}1, 8\texttt{]},\texttt{[}9, 16\texttt{]}                           & 2 & 4.41 ($-0.28$)\\
\texttt{[}1, 10\texttt{]},\texttt{[}11, 16\texttt{]}                         & 2 & 4.49 ($-0.20$)\\
\texttt{[}1, 12\texttt{]},\texttt{[}13, 16\texttt{]}                         & 2 & 4.56 ($-0.13$)\\
\midrule
\rowcolor{gray!30}
\texttt{[}1, 4\texttt{]},\texttt{[}5, 8\texttt{]},\texttt{[}9, 16\texttt{]}  & 3 & 4.23 ($-0.46$)\\
\texttt{[}1, 6\texttt{]},\texttt{[}7, 12\texttt{]},\texttt{[}13, 16\texttt{]}& 3 & 4.33 ($-0.36$)\\
\texttt{[}1, 7\texttt{]},\texttt{[}8, 11\texttt{]},\texttt{[}12, 16\texttt{]}& 3 & 4.36 ($-0.33$)\\
\texttt{[}1, 8\texttt{]},\texttt{[}9, 12\texttt{]},\texttt{[}13, 16\texttt{]}& 3 & 4.41 ($-0.28$)\\
\midrule
\texttt{[}1, 2\texttt{]},\texttt{[}3, 6\texttt{]},\texttt{[}7, 12\texttt{]},\texttt{[}13, 16\texttt{]}& 4 & 4.26 ($-0.43$)\\
\texttt{[}1, 6\texttt{]},\texttt{[}7, 12\texttt{]},\texttt{[}13, 14\texttt{]},\texttt{[}15, 16\texttt{]}& 4 & 4.32 ($-0.37$)\\
\bottomrule
\end{tabular}
\vspace{-6pt}
\end{table}

\subsection{Mask Guided Height Sampling}
\label{sub:Mask Guided Height Sampling}
As shown in the green part of Fig.~\ref{fig:model}, our MGHS generates depth-based feature $\boldsymbol{F}_\mathrm{db}$ and height-refined feature $\boldsymbol{F}_\mathrm{hr}$, respectively.
For $\boldsymbol{F}_\mathrm{db}$, we directly employ BEVPooling to project $\boldsymbol{F}_\mathrm{ctx}$.
To further refine $\boldsymbol{F}_\mathrm{db}$, we focus on utilizing informative local height ranges to produce $\boldsymbol{F}_\mathrm{hr}$, thereby addressing the loss of height information in $\boldsymbol{F}_\mathrm{db}$. In the following, we elaborate on the two core processes, named Height Decoulping and Mask Projection.
For clarity, all references to height in the following text pertain specifically to the height in voxel space.


\textbf{Height Decoupling}.
First, we conduct a comprehensive analysis of the dataset.
As shown in Fig. \ref{fig:heatmap_CDF} (a), classes such as \emph{drivable surface}, \emph{sidewalk} and \emph{terrain} exhibit similar height distributions, with the majority concentrated within the 1 to 4 height range. Conversely, categories like \emph{trailer}, \emph{manmade} and \emph{vegetation} display a broader distribution across various heights.
From a geometric perspective, the cumulative distribution function (CDF) curve in Fig. \ref{fig:heatmap_CDF} (b) reveals that the distribution deviates from either normal or uniform, with high density observed in the lower regions.

Based on the observations above, we first decouple height into different intervals $I=\{[1, 4],[5, 8],[9, 16]\}$, and then decompose features across height intervals to obtain three subspaces (L, M, and H) with distinct semantic information.
Fig. \ref{fig:Height_Sep3_with_statics} provides a detailed illustration of our approach. It can be discovered that the subspace L is predominantly occupied by \emph{drivable surface}. In contrast, in subspace H, the number of categories significantly decreases, with \emph{vegetation} and \emph{manmade} becoming more prevalent.
This decoupling strategy enables more concentrated feature representation within each subspace, enhancing the precision of the projection results.

Furthermore, we present the weighted average entropy to demonstrate the effectiveness of height decoupling:
\begin{equation}
E =-\frac{1}{N_\mathrm{sam}}\sum_{k=1}^{N_\mathrm{h}} \frac{s_k}{s_\mathrm{vox}}(\sum_{j=1}^{N_\mathrm{cla}}\frac{q_j}{N_\mathrm{vox}} \log _2 \frac{q_j}{N_\mathrm{vox}}),
\end{equation}
where $E$ and $N_\mathrm{sam}$ denote the entropy value and the total number of samples, respectively. $N_\mathrm{h}$ is the number of height intervals. $s_k$ and $s_\mathrm{vox}$ represent the size of the subspace for $k$-th interval and the total voxel grid, respectively. $q_j$ refers to the count of class $j$. $N_\mathrm{cla}$ is the number of classes and $N_\mathrm{vox}$ represents the number of voxels.

Tab.~\ref{tab:entropy_results} lists the entropy of different decoupling cases. It is observed that the case of $\{[1, 4],[5, 8],[9, 16]\}$ achieves the most significant reduction in entropy compared with other settings.
These results indicate that the proposed height decoupling effectively captures the intrinsic structure of the data, making the information in $I$ more organized.   
\begin{figure}[t]
    \centering
    \includegraphics[width=8.0cm]{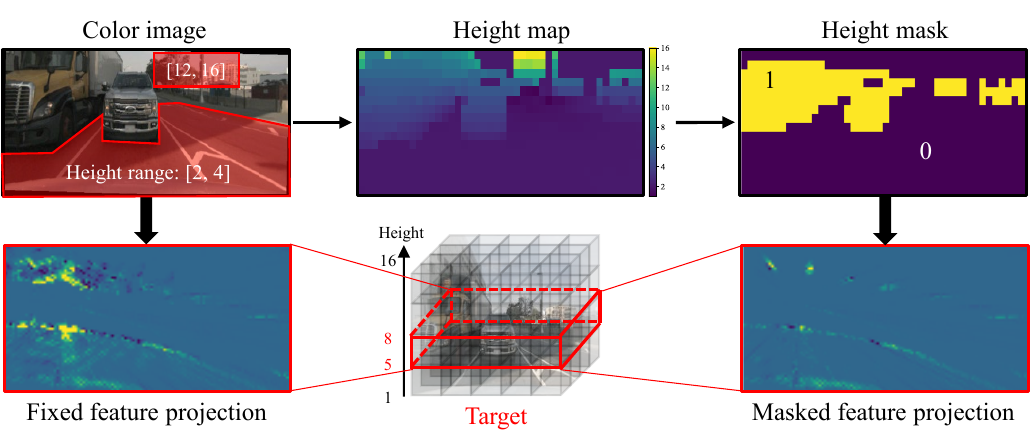}
    \vspace{-2pt}
    \caption{Comparison of projection in height range [5, 8].}
    \label{fig:mask_projection}
    \vspace{-9pt}
\end{figure}

\textbf{Mask Projection.}
To effectively capture the precise features within specific height ranges, we utilize $\boldsymbol{H}_\mathrm{mask}$ to filter out redundant feature points, generating height-aware feature $\boldsymbol{F}_{\mathrm{ha}}$.
Specifically, $\boldsymbol{H}_\mathrm{map}$ is first transformed into $\boldsymbol{H}_\mathrm{mask}$ based on $I$:
\begin{equation}
    \boldsymbol{H}_\mathrm{mask}^{k}(u,v) = 
    \begin{cases} 
    1, & \boldsymbol{H}_\mathrm{map}^{k}(u,v) \in I_k, \\
    0, & \text{otherwise,}
    \end{cases}
\end{equation}
where $k$ indexes the height interval. $\boldsymbol{F}_{\mathrm{ha}}^{k}$ is obtained as:
\begin{equation}
\boldsymbol{F}_{\mathrm{ha}}^{k} = \boldsymbol{H}_\mathrm{mask}^{k} \odot \boldsymbol{F}_{\mathrm{ctx}},
\label{Eq:mask}
\end{equation}
where $\odot$ denotes element-wise multiplication. 

Next, we project $\boldsymbol{F}_{\mathrm{ha}}$ into their respective height subspaces based on $\boldsymbol{D}$, ensuring precise spatial alignment. 
This process allows the model to capture detailed and contextually relevant information from the environment. 

Fig.~\ref{fig:mask_projection} further demonstrates the effectiveness of our mask projection. When projecting 2D features into Subspace M (height range [5, 8]) without $\boldsymbol{H}_\mathrm{mask}$, the projected features contains features from other height ranges.
However, the application of the $\boldsymbol{H}_\mathrm{mask}$ effectively filters out these extraneous features outside the specific height range, resulting in more accurate and localized projection outcomes.

\begin{table*}[t]
\centering
\Huge
\caption{3D occupancy prediction performance on the Occ3D-nuScenes dataset. The \first{best} and \second{second best} are colored}
\renewcommand\arraystretch{1.2}
\resizebox{1.0\textwidth}{!}{
\begin{tabular}{l|cc|c|c|ccccccccccccccccc|c}
\toprule
Method
& \begin{sideways}{History Frame}\end{sideways}
& \begin{sideways}{Resolution}\end{sideways}  
& \begin{sideways}{Backbone}\end{sideways}      
& \begin{sideways}{\textbf{mIoU} $\uparrow$}\end{sideways}  
& \begin{sideways}{\textcolor{otherscolor}{$\blacksquare$} \textbf{others}  $\uparrow$}\end{sideways}  
& \begin{sideways}{\textcolor{barriercolor}{$\blacksquare$} \textbf{barrier} $\uparrow$}\end{sideways}  
& \begin{sideways}{\textcolor{bicyclecolor}{$\blacksquare$} \textbf{bicycle} $\uparrow$}\end{sideways}  
& \begin{sideways}{\textcolor{buscolor}{$\blacksquare$} \textbf{bus} $\uparrow$}\end{sideways}  
& \begin{sideways}{\textcolor{carcolor}{$\blacksquare$} \textbf{car} $\uparrow$}\end{sideways}  
& \begin{sideways}{\textcolor{constructcolor}{$\blacksquare$} \textbf{const. veh.} $\uparrow$}\end{sideways}  
& \begin{sideways}{\textcolor{motorcolor}{$\blacksquare$} \textbf{motorcycle} $\uparrow$}\end{sideways}  
& \begin{sideways}{\textcolor{pedestriancolor}{$\blacksquare$} \textbf{pedestrian} $\uparrow$}\end{sideways}  
& \begin{sideways}{\textcolor{trafficcolor}{$\blacksquare$} \textbf{traffic cone} $\uparrow$}\end{sideways}  
& \begin{sideways}{\textcolor{trailercolor}{$\blacksquare$} \textbf{trailer} $\uparrow$}\end{sideways}  
& \begin{sideways}{\textcolor{truckcolor}{$\blacksquare$} \textbf{truck} $\uparrow$}\end{sideways}  
& \begin{sideways}{\textcolor{drivablecolor}{$\blacksquare$} \textbf{drive. surf.} $\uparrow$}\end{sideways}  
& \begin{sideways}{\textcolor{otherflatcolor}{$\blacksquare$} \textbf{other flat} $\uparrow$}\end{sideways}  
& \begin{sideways}{\textcolor{sidewalkcolor}{$\blacksquare$} \textbf{sidewalk} $\uparrow$}\end{sideways}  
& \begin{sideways}{\textcolor{terraincolor}{$\blacksquare$} \textbf{terrain} $\uparrow$}\end{sideways}  
& \begin{sideways}{\textcolor{manmadecolor}{$\blacksquare$} \textbf{manmade} $\uparrow$}\end{sideways}  
& \begin{sideways}{\textcolor{vegetationcolor}{$\blacksquare$} \textbf{vegetation} $\uparrow$}\end{sideways}
& \begin{sideways}{Venue}\end{sideways}\\ 
\midrule
MonoScene\cite{cao2022monoscene} & \xmark & 928 × 1600 & R101 & 6.06  & 1.75  & 7.23  & 4.26  & 4.93  & 9.38  & 5.67  & 3.98  & 3.01  & 5.90  & 4.45  & 7.17  & 14.91  & 6.32  & 7.92  & 7.43  & 1.01  & 7.65  & CVPR’22  \\
CTF-Occ\cite{tian2024occ3d} & \xmark & 928 × 1600 & R101 & 28.53  & \second{8.09}  & 39.33  & \second{20.56}  & 38.29  & 42.24  & 16.93  & \first{24.52}  & \second{22.72}  & \second{21.05}  & 22.98  & 31.11  & 53.33  & 33.84  & 37.98  & 33.23  & 20.79  & 18.00  & NIPS'24  \\
TPVFormer\cite{huang2023tri} & \xmark & 928 × 1600 & R101 & 27.83  & 7.22  & 38.90  & 13.67  & \first{40.78}  & \second{45.90}  & \second{17.23}  & 19.99  & 18.85  & 14.30  & 26.69  & \first{34.17}  & 55.65  & 35.47  & 37.55  & 30.70  & 19.40  & 16.78 & CVPR’23  \\ 
OccFormer\cite{zhang2023occformer} & \xmark & 256 × 704 & R50 & 20.40 & 6.62 & 32.57 & 13.13 & 20.37 & 37.12 & 5.04 & 14.02 & 21.01 & 16.96 & 9.34 & 20.64 & 40.89 & 27.02 & 27.43 & 18.65 & 18.78 & 16.90 & ICCV’23  \\ 
BEVDetOcc\cite{huang2021bevdet} & \xmark & 256 × 704 & R50 & 31.64  & 6.65  & 36.97  & 8.33  & 38.69  & 44.46  & 15.21  & 13.67  & 16.39  & 15.27  & 27.11  & 31.04  & \second{78.70}  & 36.45  & \second{48.27}  & \second{51.68}  & \second{36.82}  & \second{32.09} & arXiv’22  \\ 
FlashOcc\cite{yu2023flashocc} & \xmark & 256 × 704 & R50 & \second{31.95}  & 6.21  & \second{39.57}  & 11.27  & 36.32  & 43.95  & 16.25  & 14.73  & 16.89  & 15.76  & \second{28.56}  & 30.91  & 78.16  & \second{37.52}  & 47.42  & 51.35  & 36.79  & 31.42  & arXiv’23 \\
\rowcolor{gray!30}
DHD-S (Ours) & \xmark & 256 × 704 & R50 & \first{36.50}  & \first{10.59}  & \first{43.21}  & \first{23.02}  & \second{40.61}  & \first{47.31}  & \first{21.68}  & \second{23.25}  & \first{23.85}  & \first{23.40}  & \first{31.75}  & \second{34.15}  & \first{80.16}  & \first{41.30}  & \first{49.95}  & \first{54.07}  & \first{38.73}  & \first{33.51}  & - \\
\midrule
FastOcc\cite{hou2024fastocc} & 16 & 256 × 704 & R101 & 39.21  & 12.06 & 43.53 & 28.04 & 44.80 & 52.16 & 22.96 & 29.14 & 29.68 & 26.98 & 30.81 & 38.44 & 82.04 & 41.93 & 51.92 & 53.71 & 41.04 & 35.49 & ICRA'24  \\
FBOcc\cite{li2023fbocc} & 16 & 256 × 704 & R50 & 39.11  & 13.57 & 44.74 & 27.01 & 45.41 & 49.10 & 25.15 & 26.33 & 27.86 & 27.79 & 32.28 & 36.75 & 80.07 & 42.76 & 51.18 & 55.13 & 42.19 & 37.53 & ICCV’23  \\
COTR(TPVFormer)\cite{ma2024cotr} & 8 & 256 × 704 & R50 & 39.30  & 11.66 & 45.47 & 25.34 & 41.71 & 50.77 & 27.39 & 26.30 & 27.76 & 29.71 & 33.04 & 37.76 & 80.52 & 41.67 & 50.82 & 54.54 & 44.91 & 38.27 & CVPR’24  \\ 
COTR(OccFormer)\cite{ma2024cotr} & 8 & 256 × 704 & R50 & 41.20 & 12.19 & 48.47 & 27.81 & 44.28 & 52.82 & 28.70 & 28.16 & 28.95 & 31.32 & 35.01 & 39.93 & 81.54 & 42.05 & 53.44 & 56.22 & 47.37 & 41.38 & CVPR’24  \\ \cline{2-22}
BEVDetOcc-4D-Stereo\cite{huang2021bevdet} & 1 & 256 × 704 & R50 & 36.01  & 8.22 & 44.21 & 10.34 & 42.08 & 49.63 & 23.37 & 17.41 & 21.49 & 19.70 & 31.33 & 37.09 & 80.13 & 37.37 & 50.41 & 54.29 & 45.56 & 39.59 & arXiv’22  \\ 
FlashOcc\cite{yu2023flashocc} & 1 & 256 × 704 & R50 & 37.84  & 9.08 & 46.32 & 17.71 & 42.70 & 50.64 & 23.72 & 20.13 & 22.34 & 24.09 & 30.26 & 37.39 & 81.68 & 40.13 & 52.34 & 56.46 & 47.69 & \second{40.60} & arXiv’23  \\ 
COTR(BEVDetOcc)\cite{ma2024cotr} & 1 & 256 × 704 & R50 &  \second{41.39}  & \second{12.20} & \second{48.51} & \first{29.08} & \second{44.66} & \second{53.33} & \second{27.01} & \second{29.19} & \second{28.91} & \first{30.98} & \second{35.03} & 39.50 & \second{81.83} & \second{42.53} & \second{53.71} & \second{56.86} & \second{48.18} & \first{42.09} & CVPR’24  \\ 
OSP\cite{shi2024occupancysetpoints} & 1 & 900 × 1600 & R101 &  39.41  & 11.20 & 47.25 & \second{27.06} & \first{47.57} & \first{53.66} & 23.21 & \first{29.37} & \first{29.68} & 28.41 & 32.39 & \second{39.94} & 79.35 & 41.36 & 50.31 & 53.23 & 40.52 & 35.39 & ECCV’24 \\
\rowcolor{gray!30}
DHD-M (Ours) & 1 & 256 × 704 & R50 &  \first{41.49}  & \first{12.72}  & \first{48.68}  & 26.31  & 43.22  & 52.92  & \first{27.33}  & 28.49  & 28.52  & \second{30.02}  & \first{35.81}  & \first{40.24}  & \first{83.12}  & \first{44.67}  & \first{54.71}  & \first{57.69}  & \first{48.87}  & \first{42.09}  & -  \\
\hline
COTR(BEVDetOcc)\cite{ma2024cotr} & 8 & 512 × 1408 & SwinB & 46.20  & 14.85 & 53.25 & 35.19 & 50.83 & 57.25 & 35.36 & 34.06 & 33.54 & 37.14 & 38.99 & 44.97 & 84.46 & 48.73 & 57.60 & 61.08 & 51.61 & 46.72 & CVPR’24  \\
GEOcc\cite{tan2024geocc} & 8 & 512 × 1408 & SwinB & 44.67  & 14.02  & 51.40  & 33.08  & 52.08  & 56.72  & 30.04  & 33.54  & 32.34  & 35.83  & 39.34  & 44.18  & 83.49  & 46.77  & 55.72  & 58.94  & 48.85  & 43.00 &  arXiv’24 \\ \cline{2-22} 
BEVDetOcc-4D-Stereo\cite{huang2021bevdet} & 1 & 512 × 1408 & SwinB & 42.02  & 12.15 & 49.63 & 25.10 & \second{52.02} & 54.46 & \second{27.87} & 27.99 & 28.94 & 27.23 & 36.43 & 42.22 & 82.31 & 43.29 & 54.62 & 57.90 & 48.61 & 43.55 & arXiv’22  \\
FlashOcc\cite{yu2023flashocc} & 1 & 512 × 1408 & SwinB & \second{43.52}  & \second{13.42} & \second{51.07} & \second{27.68} & 51.57 & \second{56.22} & 27.27 & \second{29.98} & \second{29.93} & \second{29.80} & \second{37.77} & \second{43.52} & \second{83.81} & \second{46.55} & \second{56.15} & \second{59.56} & \second{50.84} & \second{44.67} & arXiv’23  \\ 
\rowcolor{gray!30}
DHD-L (Ours) & 1 & 512 × 1408 & SwinB & \first{45.53}  & \first{14.08} & \first{53.12} & \first{32.39} & \first{52.44} & \first{57.35} & \first{30.83} & \first{35.24} & \first{33.01} & \first{33.43} & \first{37.90} & \first{45.34} & \first{84.61} & \first{47.96} & \first{57.39} & \first{60.32} & \first{52.27} & \first{46.24} & -  \\ 
\bottomrule
\end{tabular}}
\label{tab:nuscenes_res}
\end{table*}

\begin{figure}[t]
    \centering
    \includegraphics[width=7.8cm]{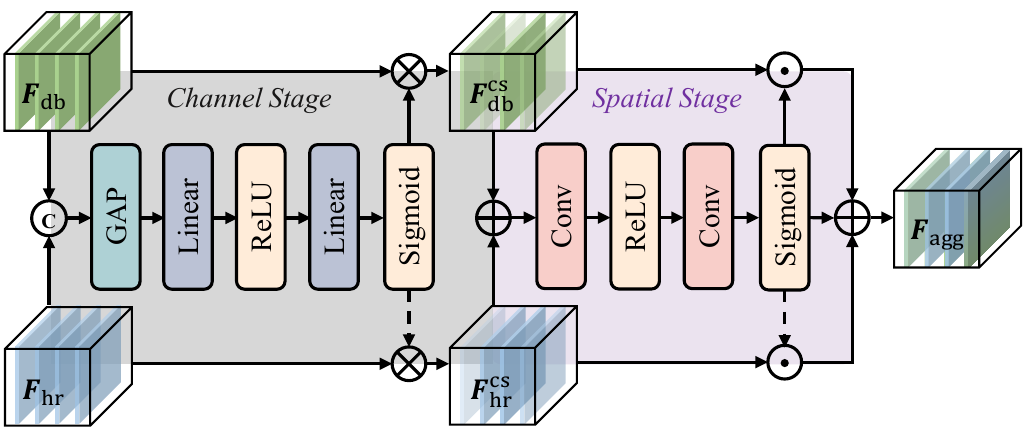}
    \caption{The proposed synergistic feature aggregation (SFA) module.}
    \label{fig:mix}
    \vspace{-7.8pt}
\end{figure}
\subsection{Synergistic Feature Aggregation}
\label{sub:Synergistic Feature Aggregation}

The key idea of the aggregation module is to select and construct the most relevant features for the occupancy prediction through a two-stage attention mechanism. Our SFA is shown in Fig.~\ref{fig:mix}. The depth-based feature $\boldsymbol{F}_\mathrm{db}$ and height-refined feature $\boldsymbol{F}_\mathrm{hr}$ are first input into the channel stage, producing the affinity vector $\boldsymbol{a}_1$:

\begin{equation}
\boldsymbol{a}_1=\sigma(\boldsymbol{W}_2(\delta(\boldsymbol{W}_1(\boldsymbol{f}_\mathrm{gap}(\boldsymbol{f}_\mathrm{cat}(\boldsymbol{F}_\mathrm{db},\boldsymbol{F}_\mathrm{hr})))))),
\end{equation}
where $\sigma(\cdot)$, $\delta(\cdot)$, $f_\mathrm{gap}(\cdot)$, and $f_\mathrm{cat}(\cdot)$ refer to the sigmoid, ReLU, global average pooling, and concatenation, respectively. 
$\boldsymbol{W}_1$ and $\boldsymbol{W}_2$ denote linear layers.
Then, $\boldsymbol{a}_1$ is employed to adaptively weight $\boldsymbol{F}_\mathrm{db}$ and $\boldsymbol{F}_\mathrm{hr}$, resulting in the channel-enhanced features $\boldsymbol{F}_\mathrm{db}^\mathrm{cs}$ and $\boldsymbol{F}_\mathrm{hr}^\mathrm{cs} $, where $\boldsymbol{F}_\mathrm{db}^\mathrm{cs}=\boldsymbol{a}_1\boldsymbol{F}_\mathrm{db}$ and $\boldsymbol{F}_\mathrm{hr}^\mathrm{cs}=(1-\boldsymbol{a}_1)\boldsymbol{F}_\mathrm{hr}$.

Next, the spatial stage takes $\boldsymbol{F}_\mathrm{db}^\mathrm{cs}$ and $\boldsymbol{F}_\mathrm{hr}^\mathrm{cs}$ as input to obtain the spatial affinity map $\boldsymbol{A}_2$:
\begin{equation}
\boldsymbol{A}_2=\sigma( 
f_\mathrm{conv}(
    \delta(
        f_\mathrm{conv}(\boldsymbol{F}_\mathrm{db}^\mathrm{cs} + \boldsymbol{F}_\mathrm{hr}^\mathrm{cs}))) ),
\end{equation}
where $f_\mathrm{conv}(\cdot)$ is convolution layer.
Finally, $\boldsymbol{A}_2$ is used to further enhance the spatial information of the channel stage and dynamically aggregate the channel-enhanced features:
\begin{equation}
\boldsymbol{F}_\mathrm{agg}=\boldsymbol{A}_2 \odot \boldsymbol{F}_\mathrm{db}^\mathrm{cs} + (1-\boldsymbol{A}_2) \odot \boldsymbol{F}_\mathrm{hr}^\mathrm{cs},
\end{equation}
where $\boldsymbol{F}_\mathrm{agg}$ represents the output features of SFA.

\subsection{Training Loss}
\label{sub:training loss}

First, we utilize binary cross-entropy loss $\mathcal{L}_{\mathrm{bce}}$ to constrain the training of DepthNet and HeightNet:
\begin{equation}
    \mathcal{L}_{\mathrm{bce}} = - \sum_{g=1}^{N_\mathrm{gt}} \left( \hat{p}_{g} \log(p_{g}) + (1 - \hat{p}_{g}) \log(1 - p_{g}) \right),
\end{equation}
where $N_\mathrm{gt}$ is the number of ground-truth pixels, $p_{g}$ and $\hat{p}_{g}$ are the prediction and label of $g$-th pixel, respectively.

Then, a weighted cross-entropy loss $\mathcal{L}_{\mathrm{ce}}$ is used to supervise the learning of the predictor:
\begin{equation}
    \mathcal{L}_{\mathrm{ce}}=-\sum_{i=1}^{N_\mathrm{vox}} \sum_{j=1}^{N_\mathrm{cla}} w_j \hat{r}_{i, j} \log \left(\frac{e^{r_{i, j}}}{\sum_j e^{r_{i, j}}}\right),
\end{equation}
where $N_\mathrm{vox}$ and $N_\mathrm{cla}$ represent the total number of the voxels and classes. $r_{i,j}$ is the prediction for $i$-th voxel belonging to class $j$, $\hat{r}_{i, j}$ is the corresponding label.
$w_j$ is a weight for each according to the inverse of the class frequency. Additionally, inspired by MonoScene~\cite{cao2022monoscene}, $\mathcal{L}_\mathrm{scal}^\mathrm{sem}$ and $\mathcal{L}_\mathrm{scal}^\mathrm{geo}$ are introduced to constrain the learning of our method.

Finally, the total training loss $\mathcal{L}$ is defined as:
\begin{equation}
\mathcal{L}=\lambda_1 \mathcal{L}_\mathrm{bce}^\mathrm{depth}+\lambda_2 \mathcal{L}_\mathrm{bce}^\mathrm{height}+\lambda_3 \mathcal{L}_\mathrm{ce}+\lambda_4 \mathcal{L}_\mathrm{scal}^\mathrm{sem}+\lambda_5 \mathcal{L}_\mathrm{scal}^\mathrm{geo},
\label{Eq:loss}
\end{equation}
where $\lambda_1$, $\lambda_2$, $\lambda_3$, $\lambda_4$, and $\lambda_5$ are hyper-parameters.
Note that in DHD-S, depth supervision is not utilized.

\section{Experiment}
\subsection{Datasets}
We implement our DHD on the large-scale Occ3D-nuScenes~\cite{tian2024occ3d} dataset, including 700 scenes for training and 150 for validation. 
It spans a spatial range of -40m to 40m along both the X and Y axes, and -1m to 5.4m along the Z axis. 
Besides, the occupancy labels are defined using voxels with a size of $0.4\rm m \times 0.4\rm m \times 0.4\rm m$, containing 17 distinct categories.
Each driving scene encapsulates 20 seconds of annotated data, captured at a frequency of 2Hz. 

\subsection{Implementation Details}
We present three versions of DHD, namely DHD-S, DHD-M, and DHD-L. Specifically, DHD-S employs ResNet50 \cite{he2016deep} as the image backbone, without incorporating temporal information. DHD-M integrates BEVStereo \cite{li2023bevstereo} for view transformation.
Building upon DHD-M, our DHD-L replaces the image backbone with SwinTransformer-B \cite{liu2021swin} and increases the image resolution to $512 \times 1408$. 

Following previous methods~\cite{yu2023flashocc, ma2024cotr, shi2024occupancysetpoints}, the mean intersection over union (mIoU) is employed as the evaluation metric. During training, we utilize the AdamW optimizer~\cite{loshchilov2017decoupled} with a learning rate of $2\times 10^{-4} $ to train our DHD for 24 epochs. The proposed model is implemented based on MMDetection3D~\cite{mmdet3d2020} with six GeForce RTX 4090 GPUs. Additionally, the hyper-parameters are set as $\lambda_1=0.05$, $\lambda_2=0.1$, $\lambda_3=10$, and $\lambda_4=\lambda_5=0.2$.


\subsection{Comparison with the State-of-the-Art}
We compare DHD-S, DHD-M, and DHD-L with state-of-the-art methods on the Occ3D-nuScenes \cite{tian2024occ3d} dataset. 
To ensure fairness, we categorize and compare existing methods based on temporal information, resolution, and backbone.

\textbf{Quantitative Comparison.} Tab.~\ref{tab:nuscenes_res} demonstrates that our DHD achieves state-of-the-art performance on the Occ3D-nuScenes dataset. For the ResNet50~\cite{he2016deep} backbone and the input resolution of $256\times704$, compared to the second-best method (FlashOcc~\cite{yu2023flashocc}) without history frames, our DHD-S increases the mIoU by 4.55. Additionally, when a single history frame is introduced, our DHD-M surpasses the suboptimal method (COTR~\cite{ma2024cotr}) by 0.10 in mIoU. For the SwinTransformer-B~\cite{liu2021swin} backbone and the input resolution of $512\times1408$, our DHD-L obtains the best performance in the single history frame setting. For example, our DHD-L outperforms the FlashOcc~\cite{yu2023flashocc} by 2.01 in mIoU and even surpassing GEOcc~\cite{tan2024geocc}, which employs long-term temporal information, by 0.86 mIoU.
In short, these quantitative results indicate that our method significantly improves prediction accuracy even with minimal input frames.

\begin{figure*}[t]
    \centering
    \includegraphics[width=17.7cm]{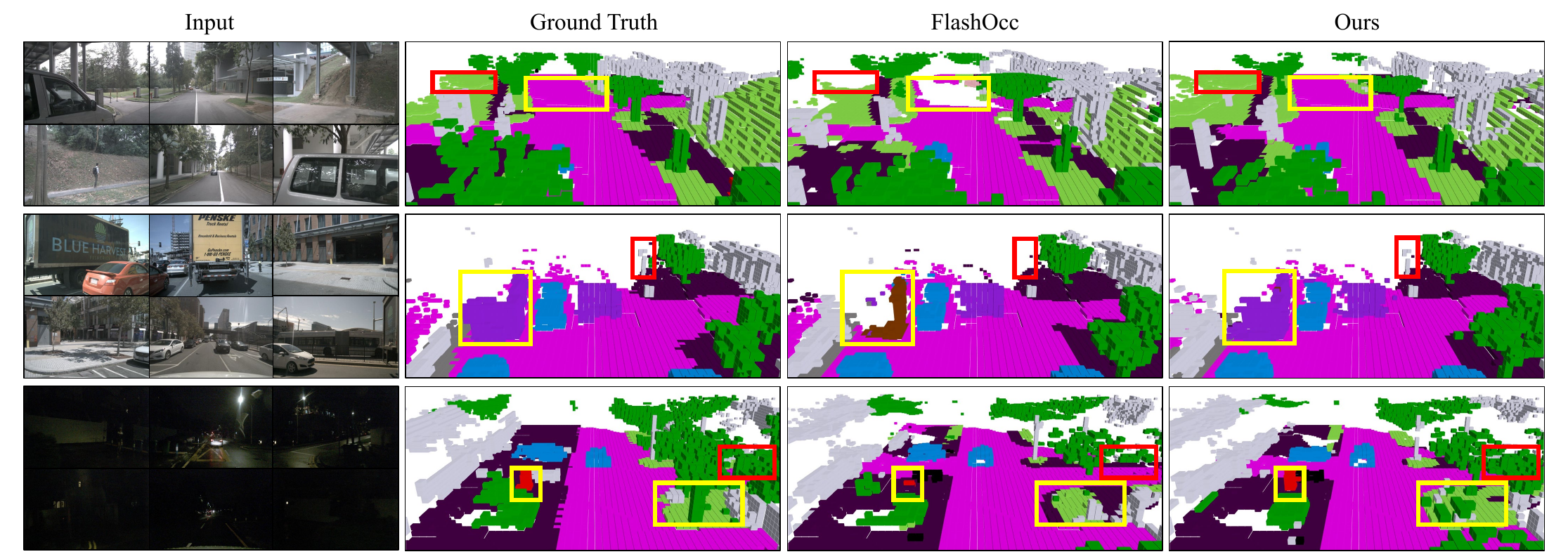}
    \vspace{-20pt}
    \caption{Qualitative comparison of our method and FlashOcc \cite{yu2023flashocc} on the Occ3D-nuScenes dataset.}
    \label{fig:vis}
\end{figure*}

\textbf{Visual Comparison.} Fig.~\ref{fig:vis} presents the visual comparison of DHD and FlashOcc~\cite{yu2023flashocc} configured with a resolution of $512 \times 1408$. Notably, our method can predict more accurate occupancy results. For example, compared to FlashOcc~\cite{yu2023flashocc}, the first row demonstrates that our DHD performs better on \emph{drive. surf.} and \emph{terrain}, both of which are primarily concentrated at lower heights.
This improvement highlights DHD's ability to effectively handle lower-level features.
The second and third rows show that our model excels in differentiating objects within similar height ranges, such as \emph{truck} and \emph{trailer}. 
Additionally, DHD accurately estimate the instance of \emph{pedestrian} and \emph{vegetation}.
These visual results demonstrate that our approach achieves a more precise and comprehensive understanding of the 3D scene. 

\begin{table}[t]
\centering
\caption{Ablation study of MGHS and SFA.}
\vspace{-5pt}
\label{tab:ablation}
\resizebox{0.45\textwidth}{!}{%
\renewcommand{\arraystretch}{1.2}
\begin{tabular}{c|cc|c|c}
\toprule
\multirow{2}{*}{DHD-S} & \multicolumn{2}{c|}{MGHS} & \multirow{2}{*}{SFA} & \multirow{2}{*}{mIoU} \\ \cline{2-3}
    & Height Decoupling & Mask Projection  &        &       \\ 
\midrule
i   &           &            &          & 33.72 \\
ii  & \cmark    &            &          & 35.15 \\
iii & \cmark    & \cmark     &          & 35.60  \\
iv  & \cmark    & \cmark     & \cmark   &\textbf{35.90}  \\ 
\bottomrule
\end{tabular}%
}
\end{table}

\subsection{Ablation Study}
For fast validation, all ablations are deployed on the small model DHD-S, where only the cross-entropy loss is used. 

\textbf{MGHS and SFA.} Tab.~\ref{tab:ablation} reports the ablation study of MGHS and SFA on the Occ3D-nuScenes dataset. The baseline (i)  deletes the SFA and removes the height decoupling part in MGHS while preserving the depth-based feature $\boldsymbol{F}_\mathrm{db}$. (ii) and (iii) replace the SFA with concatenation, indicating that both the height decoupling and mask projection in MGHS contribute to mIoU improvements. When the SFA is employed on the basis of (iii), (iv) achieves the best performance. For example, compared to the baseline (i), the model (iv) increases the mIoU by 2.18. These results indicate that our MGHS and SFA can significantly improve the accuracy of occupancy prediction. 

\begin{figure}[t]
    \centering
    \includegraphics[width=8.6cm]{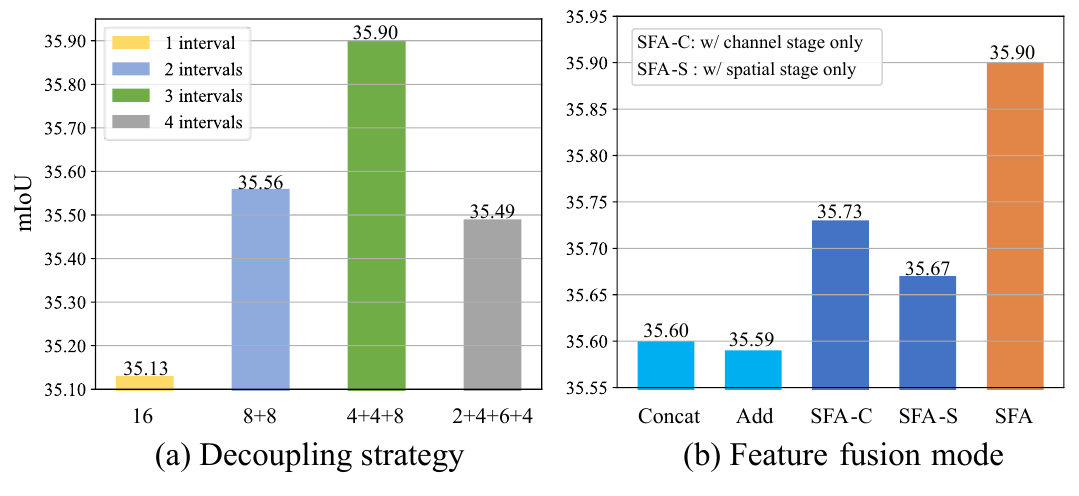}
    \vspace{-16pt}
    \caption{Ablation study of (a) different decoupling strategies and (b) feature fusion modes. 16: [1, 16], 8 + 8: \{[1, 8], [9, 16]\}, 4 + 4 + 8: \{[1, 4], [5, 8], [9, 16]\}, 2 + 4 + 6 + 4: \{[1, 2], [3, 6], [7, 12], [13, 16]\}.}
    \label{fig:ablation_MGHS_mix}
\end{figure}

\textbf{Height Decoupling.} Fig. \ref{fig:ablation_MGHS_mix} (a) presents the ablation of different height decoupling strategies. The baseline (yellow bar) employs SFA and MGHS with [1, 16] height decoupling. We find that decoupling the height into multiple intervals brings consistent improvement over the single interval. Notably, when implementing the `4+4+8' decoupling, the model (green bar) achieves the best performance, surpassing the baseline by 0.77 in mIoU. These results validate the rationale behind the height decoupling strategy, as determined by the data statistics presented in Fig.~\ref{fig:box} and Fig.~\ref{fig:heatmap_CDF}, as well as the data analysis in Tab.~\ref{tab:entropy_results}.

\textbf{Feature Fusion.} Fig. \ref{fig:ablation_MGHS_mix} (b) illustrates the ablation study of different feature fusion modes. The baseline `Concat' corresponds to DHD-iii in Tab.~\ref{tab:ablation}. Compared to the commonly used addition and concatenation, both the channel stage (SFA-C) and the spatial stage (SFA-S) can increase the mIoU. When SFA-C and SFA-S are deployed together, the model (orange bar) achieves the best result, exceeding the addition by 0.31 and concatenation by 0.30 in mIoU. 

\section{CONCLUSIONS}
In this paper, we proposed the novel deep height decoupling (DHD) framework for vision-based 3D occupancy prediction. For the first time, DHD introduced the explicit height prior to filter the confusing occupancy features. 
Specifically, DHD estimated the height map and decoupled it into multiple binary masks based on reliable height distribution statistics, including heatmap visualization, cumulative distribution function curve, and entropy calculation. 
Then the mask guided height sampling was designed to realize the more accurate 2D-to-3D view transformation. In addition, at the end of the model, the two-stage synergistic feature aggregation is introduced to enhance the feature representation using channel and spatial affinities. 
Owing to these designs, DHD achieved state-of-the-art performance even with minimal input frames on Occ3D-nuScenes benchmark.



\section*{ACKNOWLEDGMENT}
This work was supported by the National Science Fund of China under Grant Nos. U24A20330 and 62361166670.

\newpage
\bibliographystyle{IEEEtran}
\bibliography{ref}

\begin{thebibliography}{10}
\providecommand{\url}[1]{#1}
\csname url@samestyle\endcsname
\providecommand{\newblock}{\relax}
\providecommand{\bibinfo}[2]{#2}
\providecommand{\BIBentrySTDinterwordspacing}{\spaceskip=0pt\relax}
\providecommand{\BIBentryALTinterwordstretchfactor}{4}
\providecommand{\BIBentryALTinterwordspacing}{\spaceskip=\fontdimen2\font plus
\BIBentryALTinterwordstretchfactor\fontdimen3\font minus \fontdimen4\font\relax}
\providecommand{\BIBforeignlanguage}[2]{{%
\expandafter\ifx\csname l@#1\endcsname\relax
\typeout{** WARNING: IEEEtran.bst: No hyphenation pattern has been}%
\typeout{** loaded for the language `#1'. Using the pattern for}%
\typeout{** the default language instead.}%
\else
\language=\csname l@#1\endcsname
\fi
#2}}
\providecommand{\BIBdecl}{\relax}
\BIBdecl

\bibitem{huang2021bevdet}
J.~Huang, G.~Huang, Z.~Zhu, Y.~Ye, and D.~Du, ``Bevdet: High-performance multi-camera 3d object detection in bird-eye-view,'' \emph{arXiv preprint arXiv:2112.11790}, 2021.

\bibitem{li2022bevformer}
Z.~Li, W.~Wang, H.~Li, E.~Xie, C.~Sima, T.~Lu, Y.~Qiao, and J.~Dai, ``Bevformer: Learning bird’s-eye-view representation from multi-camera images via spatiotemporal transformers,'' in \emph{ECCV}, 2022, pp. 1--18.

\bibitem{CGFormer}
Z.~Yu, R.~Zhang, J.~Ying, J.~Yu, X.~Hu, L.~Luo, S.-Y. Cao, and H.-L. Shen, ``Context and geometry aware voxel transformer for semantic scene completion,'' in \emph{NIPS}, 2024.

\bibitem{huang2024gaussianformer}
Y.~Huang, W.~Zheng, Y.~Zhang, J.~Zhou, and J.~Lu, ``Gaussianformer: Scene as gaussians for vision-based 3d semantic occupancy prediction,'' \emph{arXiv preprint arXiv:2405.17429}, 2024.

\bibitem{peng2024learning}
L.~Peng, J.~Xu, H.~Cheng, Z.~Yang, X.~Wu, W.~Qian, W.~Wang, B.~Wu, and D.~Cai, ``Learning occupancy for monocular 3d object detection,'' in \emph{CVPR}, 2024, pp. 10\,281--10\,292.

\bibitem{yan2024tri}
Z.~Yan, Y.~Lin, K.~Wang, Y.~Zheng, Y.~Wang, Z.~Zhang, J.~Li, and J.~Yang, ``Tri-perspective view decomposition for geometry-aware depth completion,'' in \emph{CVPR}, 2024, pp. 4874--4884.

\bibitem{wang2024dcdepth}
K.~Wang, Z.~Yan, J.~Fan, W.~Zhu, X.~Li, J.~Li, and J.~Yang, ``Dcdepth: Progressive monocular depth estimation in discrete cosine domain,'' \emph{arXiv preprint arXiv:2410.14980}, 2024.

\bibitem{yan2022rignet}
Z.~Yan, K.~Wang, X.~Li, Z.~Zhang, J.~Li, and J.~Yang, ``Rignet: Repetitive image guided network for depth completion,'' in \emph{ECCV}, 2022, pp. 214--230.

\bibitem{zhu2023curricular}
Z.~Zhu, Q.~Meng, X.~Wang, K.~Wang, L.~Yan, and J.~Yang, ``Curricular object manipulation in lidar-based object detection,'' in \emph{CVPR}, 2023, pp. 1125--1135.

\bibitem{zheng2023occworld}
W.~Zheng, W.~Chen, Y.~Huang, B.~Zhang, Y.~Duan, and J.~Lu, ``Occworld: Learning a 3d occupancy world model for autonomous driving,'' \emph{arXiv preprint arXiv: 2311.16038}, 2023.

\bibitem{li2023voxformer}
Y.~Li, Z.~Yu, C.~Choy, C.~Xiao, J.~M. Alvarez, S.~Fidler, C.~Feng, and A.~Anandkumar, ``Voxformer: Sparse voxel transformer for camera-based 3d semantic scene completion,'' in \emph{CVPR}, 2023, pp. 9087--9098.

\bibitem{wei2023surroundocc}
Y.~Wei, L.~Zhao, W.~Zheng, Z.~Zhu, J.~Zhou, and J.~Lu, ``Surroundocc: Multi-camera 3d occupancy prediction for autonomous driving,'' in \emph{ICCV}, 2023, pp. 21\,729--21\,740.

\bibitem{tian2024occ3d}
X.~Tian, T.~Jiang, L.~Yun, Y.~Mao, H.~Yang, Y.~Wang, Y.~Wang, and H.~Zhao, ``Occ3d: A large-scale 3d occupancy prediction benchmark for autonomous driving,'' \emph{NIPS}, 2024.

\bibitem{gan2024gaussianocc}
W.~Gan, F.~Liu, H.~Xu, N.~Mo, and N.~Yokoya, ``Gaussianocc: Fully self-supervised and efficient 3d occupancy estimation with gaussian splatting,'' \emph{arXiv preprint arXiv:2408.11447}, 2024.

\bibitem{shi2024occupancysetpoints}
Y.~Shi, T.~Cheng, Q.~Zhang, W.~Liu, and X.~Wang, ``Occupancy as set of points,'' in \emph{ECCV}, 2024.

\bibitem{li2024viewformer}
J.~Li, X.~He, C.~Zhou, X.~Cheng, Y.~Wen, and D.~Zhang, ``Viewformer: Exploring spatiotemporal modeling for multi-view 3d occupancy perception via view-guided transformers,'' \emph{arXiv preprint arXiv:2405.04299}, 2024.

\bibitem{li2023bevdepth}
Y.~Li, Z.~Ge, G.~Yu, J.~Yang, Z.~Wang, Y.~Shi, J.~Sun, and Z.~Li, ``Bevdepth: Acquisition of reliable depth for multi-view 3d object detection,'' in \emph{AAAI}, 2023, pp. 1477--1485.

\bibitem{li2023bevstereo}
Y.~Li, H.~Bao, Z.~Ge, J.~Yang, J.~Sun, and Z.~Li, ``Bevstereo: Enhancing depth estimation in multi-view 3d object detection with temporal stereo,'' in \emph{AAAI}, 2023, pp. 1486--1494.

\bibitem{hou2024fastocc}
J.~Hou, X.~Li, W.~Guan, G.~Zhang, D.~Feng, Y.~Du, X.~Xue, and J.~Pu, ``Fastocc: Accelerating 3d occupancy prediction by fusing the 2d bird’s-eye view and perspective view,'' \emph{ICRA}, 2024.

\bibitem{yu2023flashocc}
Z.~Yu, C.~Shu, J.~Deng, K.~Lu, Z.~Liu, J.~Yu, D.~Yang, H.~Li, and Y.~Chen, ``Flashocc: Fast and memory-efficient occupancy prediction via channel-to-height plugin,'' \emph{arXiv preprint arXiv:2311.12058}, 2023.

\bibitem{yu2024panoptic}
Z.~Yu, C.~Shu, Q.~Sun, J.~Linghu, X.~Wei, J.~Yu, Z.~Liu, D.~Yang, H.~Li, and Y.~Chen, ``Panoptic-flashocc: An efficient baseline to marry semantic occupancy with panoptic via instance center,'' \emph{arXiv preprint arXiv:2406.10527}, 2024.

\bibitem{myeongjin2023milo}
T.~V. J.-H.~K. Myeongjin and K.~S. J. S.-G. Jeong, ``Milo: Multi-task learning with localization ambiguity suppression for occupancy prediction cvpr 2023 occupancy challenge report,'' \emph{arXiv preprint arXiv:2306.11414}, 2023.

\bibitem{yan2023distortion}
Z.~Yan, X.~Li, K.~Wang, S.~Chen, J.~Li, and J.~Yang, ``Distortion and uncertainty aware loss for panoramic depth completion,'' in \emph{ICML}.\hskip 1em plus 0.5em minus 0.4em\relax PMLR, 2023, pp. 39\,099--39\,109.

\bibitem{yan2022multi}
Z.~Yan, X.~Li, K.~Wang, Z.~Zhang, J.~Li, and J.~Yang, ``Multi-modal masked pre-training for monocular panoramic depth completion,'' in \emph{ECCV}.\hskip 1em plus 0.5em minus 0.4em\relax Springer, 2022, pp. 378--395.

\bibitem{cao2022monoscene}
A.-Q. Cao and R.~De~Charette, ``Monoscene: Monocular 3d semantic scene completion,'' in \emph{CVPR}, 2022, pp. 3991--4001.

\bibitem{philion2020lift}
J.~Philion and S.~Fidler, ``Lift, splat, shoot: Encoding images from arbitrary camera rigs by implicitly unprojecting to 3d,'' in \emph{ECCV}, 2020, pp. 194--210.

\bibitem{huang2022bevdet4d}
J.~Huang and G.~Huang, ``Bevdet4d: Exploit temporal cues in multi-camera 3d object detection,'' \emph{arXiv preprint arXiv:2203.17054}, 2022.

\bibitem{liu2023fully}
H.~Liu, H.~Wang, Y.~Chen, Z.~Yang, J.~Zeng, L.~Chen, and L.~Wang, ``Fully sparse 3d panoptic occupancy prediction,'' \emph{arXiv preprint arXiv:2312.17118}, 2023.

\bibitem{huang2024selfocc}
Y.~Huang, W.~Zheng, B.~Zhang, J.~Zhou, and J.~Lu, ``Selfocc: Self-supervised vision-based 3d occupancy prediction,'' in \emph{CVPR}, 2024, pp. 19\,946--19\,956.

\bibitem{zhang2023occnerf}
C.~Zhang, J.~Yan, Y.~Wei, J.~Li, L.~Liu, Y.~Tang, Y.~Duan, and J.~Lu, ``Occnerf: Self-supervised multi-camera occupancy prediction with neural radiance fields,'' \emph{arXiv preprint arXiv:2312.09243}, 2023.

\bibitem{pan2023renderocc}
M.~Pan, J.~Liu, R.~Zhang, P.~Huang, X.~Li, L.~Liu, and S.~Zhang, ``Renderocc: Vision-centric 3d occupancy prediction with 2d rendering supervision,'' \emph{arXiv preprint arXiv:2309.09502}, 2023.

\bibitem{yan2023desnet}
Z.~Yan, K.~Wang, X.~Li, Z.~Zhang, J.~Li, and J.~Yang, ``Desnet: Decomposed scale-consistent network for unsupervised depth completion,'' in \emph{AAAI}, vol.~37, no.~3, 2023, pp. 3109--3117.

\bibitem{yan2024learnable}
Z.~Yan, Y.~Zheng, D.-P. Fan, X.~Li, J.~Li, and J.~Yang, ``Learnable differencing center for nighttime depth perception,'' \emph{Visual Intelligence}, vol.~2, no.~1, p.~15, 2024.

\bibitem{wang2021regularizing}
K.~Wang, Z.~Zhang, Z.~Yan, X.~Li, B.~Xu, J.~Li, and J.~Yang, ``Regularizing nighttime weirdness: Efficient self-supervised monocular depth estimation in the dark,'' in \emph{ICCV}, 2021, pp. 16\,055--16\,064.

\bibitem{yan2024completion}
Z.~Yan, Z.~Wang, K.~Wang, J.~Li, and J.~Yang, ``Completion as enhancement: A degradation-aware selective image guided network for depth completion,'' \emph{arXiv preprint arXiv:2412.19225}, 2024.

\bibitem{yan2023rignet++}
Z.~Yan, X.~Li, L.~Hui, Z.~Zhang, J.~Li, and J.~Yang, ``Rignet++: Semantic assisted repetitive image guided network for depth completion,'' \emph{arXiv preprint arXiv:2309.00655}, 2023.

\bibitem{miao2023occdepth}
R.~Miao, W.~Liu, M.~Chen, Z.~Gong, W.~Xu, C.~Hu, and S.~Zhou, ``Occdepth: A depth-aware method for 3d semantic scene completion,'' \emph{arXiv preprint arXiv:2302.13540}, 2023.

\bibitem{zhang2023sa}
J.~Zhang, Y.~Zhang, Q.~Liu, and Y.~Wang, ``Sa-bev: Generating semantic-aware bird's-eye-view feature for multi-view 3d object detection,'' in \emph{ICCV}, 2023, pp. 3348--3357.

\bibitem{xie2023sparsefusion}
Y.~Xie, C.~Xu, M.-J. Rakotosaona, P.~Rim, F.~Tombari, K.~Keutzer, M.~Tomizuka, and W.~Zhan, ``Sparsefusion: Fusing multi-modal sparse representations for multi-sensor 3d object detection,'' in \emph{ICCV}, 2023, pp. 17\,591--17\,602.

\bibitem{huang2022bevpoolv2}
J.~Huang and G.~Huang, ``Bevpoolv2: A cutting-edge implementation of bevdet toward deployment,'' \emph{arXiv preprint arXiv:2211.17111}, 2022.

\bibitem{chi2023bev}
X.~Chi, J.~Liu, M.~Lu, R.~Zhang, Z.~Wang, Y.~Guo, and S.~Zhang, ``Bev-san: Accurate bev 3d object detection via slice attention networks,'' in \emph{CVPR}, 2023, pp. 17\,461--17\,470.

\bibitem{hu2018squeeze}
J.~Hu, L.~Shen, and G.~Sun, ``Squeeze-and-excitation networks,'' in \emph{CVPR}, 2018, pp. 7132--7141.

\bibitem{zhu2019deformable}
X.~Zhu, H.~Hu, S.~Lin, and J.~Dai, ``Deformable convnets v2: More deformable, better results,'' in \emph{CVPR}, 2019, pp. 9308--9316.

\bibitem{huang2023tri}
Y.~Huang, W.~Zheng, Y.~Zhang, J.~Zhou, and J.~Lu, ``Tri-perspective view for vision-based 3d semantic occupancy prediction,'' in \emph{CVPR}, 2023, pp. 9223--9232.

\bibitem{zhang2023occformer}
Y.~Zhang, Z.~Zhu, and D.~Du, ``Occformer: Dual-path transformer for vision-based 3d semantic occupancy prediction,'' in \emph{ICCV}, 2023, pp. 9433--9443.

\bibitem{li2023fbocc}
Z.~Li, Z.~Yu, D.~Austin, M.~Fang, S.~Lan, J.~Kautz, and J.~M. Alvarez, ``{FB-OCC}: {3D} occupancy prediction based on forward-backward view transformation,'' \emph{arXiv:2307.01492}, 2023.

\bibitem{ma2024cotr}
Q.~Ma, X.~Tan, Y.~Qu, L.~Ma, Z.~Zhang, and Y.~Xie, ``Cotr: Compact occupancy transformer for vision-based 3d occupancy prediction,'' in \emph{CVPR}, 2024, pp. 19\,936--19\,945.

\bibitem{tan2024geocc}
X.~Tan, W.~Wu, Z.~Zhang, C.~Fan, Y.~Peng, Z.~Zhang, Y.~Xie, and L.~Ma, ``Geocc: Geometrically enhanced 3d occupancy network with implicit-explicit depth fusion and contextual self-supervision,'' \emph{arXiv preprint arXiv:2405.10591}, 2024.

\bibitem{he2016deep}
K.~He, X.~Zhang, S.~Ren, and J.~Sun, ``Deep residual learning for image recognition,'' in \emph{CVPR}, 2016, pp. 770--778.

\bibitem{liu2021swin}
Z.~Liu, Y.~Lin, Y.~Cao, H.~Hu, Y.~Wei, Z.~Zhang, S.~Lin, and B.~Guo, ``Swin transformer: Hierarchical vision transformer using shifted windows,'' in \emph{ICCV}, 2021, pp. 10\,012--10\,022.

\bibitem{loshchilov2017decoupled}
I.~Loshchilov and F.~Hutter, ``Decoupled weight decay regularization,'' \emph{arXiv preprint arXiv:1711.05101}, 2017.

\bibitem{mmdet3d2020}
M.~Contributors, ``{MMDetection3D: OpenMMLab} next-generation platform for general {3D} object detection,'' \url{https://github.com/open-mmlab/mmdetection3d}, 2020.

\end{thebibliography}

\end{document}